\documentclass[lettersize,journal]{IEEEtran}

\usepackage{amsmath,amssymb,amsfonts}
\usepackage{algorithmic}
\usepackage{algorithm}
\usepackage{array}
\usepackage[caption=false,font=normalsize,labelfont=sf,textfont=sf]{subfig}
\usepackage{textcomp}
\usepackage{stfloats}
\usepackage{url}
\usepackage{verbatim}
\usepackage{graphicx}
\usepackage{cite}
\usepackage[hidelinks]{hyperref}
\usepackage{xcolor}
\hypersetup{colorlinks=true}

\usepackage{orcidlink} 
\usepackage{booktabs}
\usepackage{makecell}
\usepackage{bm}

\usepackage[capitalize,nameinlink,noabbrev]{cleveref}
\begin{document}

\title{Data-Driven Prediction of Seismic Intensity Distributions Featuring Hybrid Classification-Regression Models}
% Data-driven prediction of seismic intensity distributions featuring hybrid classification-regression models 
% Linear Regression Surpasses \\ Ground Motion Prediction Equations
 
% \author{IEEE Publication Technology,~\IEEEmembership{Staff,~IEEE,}
%         % <-this % stops a space
% \thanks{This paper was produced by the IEEE Publication Technology Group. They are in Piscataway, NJ.}% <-this % stops a space 
% \thanks{Manuscript received April 19, 2021; revised August 16, 2021.}}

\author{Koyu Mizutani\orcidlink{0009-0003-8974-3264}, Haruki Mitarai\orcidlink{0000-0002-8374-3648}, Kakeru Miyazaki, Soichiro Kumano\orcidlink{0000-0002-3461-3943}, Toshihiko Yamasaki\orcidlink{0000-0002-1784-2314}

% \thanks{Manuscript received April 19, 2021; revised August 16, 2021.}
\thanks{Koyu Mizutani, Haruki Mitarai, Soichiro Kumano, and Toshihiko Yamasaki are with the Department of Information and Communication Engineering, Graduate School of Information Science and Technology, The University of Tokyo, Tokyo, Japan~(email: mizutani@cvm.t.u-tokyo.ac.jp; hmitarai@tkl.iis.u-tokyo.ac.jp; kumano@cvm.t.u-tokyo.ac.jp; yamasaki@cvm.t.u-tokyo.ac.jp)}
\thanks{Kakeru Miyazaki is with the Emerging Design and Informatics Course, Graduate School of Interdisciplinary Information Studies, The University of Tokyo, Tokyo, Japan~(email: kakeru-miyazaki@iis-lab.org)}
}

% The paper headers
% \markboth{IEEE TRANSACTIONS ON CONSUMER ELECTRONICS}%
% {Shell \MakeLowercase{\textit{et al.}}: A Sample Article Using IEEEtran.cls for IEEE Journals}

% \IEEEpubid{0000--0000/00\$00.00~\copyright~2021 IEEE}
% Remember, if you use this you must call \IEEEpubidadjcol in the second
% column for its text to clear the IEEEpubid mark.

\maketitle

\begin{abstract}
Earthquakes are among the most immediate and deadly natural disasters that humans face. 
Accurately forecasting the extent of earthquake damage and assessing potential risks can be instrumental in saving numerous lives. 
In this study, we developed linear regression models capable of predicting seismic intensity distributions based on earthquake parameters: location, depth, and magnitude. 
Because it is completely data-driven, it can predict intensity distributions without geographical information.
The dataset comprises seismic intensity data from earthquakes that occurred in the vicinity of Japan between 1997 and 2020, specifically containing 1,857 instances of earthquakes with a magnitude of 5.0 or greater, sourced from the Japan Meteorological Agency.
We trained both regression and classification models and combined them to take advantage of both to create a hybrid model.
The proposed model outperformed commonly used Ground Motion Prediction Equations~(GMPEs) in terms of the correlation coefficient, F1 score, and MCC. 
Furthermore, the proposed model can predict even abnormal seismic intensity distributions, a task at conventional GMPEs often struggle.
\end{abstract}


\section{Introduction}
\IEEEPARstart{A}{ century} ago, the Great Kanto Earthquake struck the heart of Japan, causing unprecedented devastation and significant loss of life and infrastructure damage.
The 2011 Great East Japan Earthquake, along with its subsequent tsunami, remains vividly imprinted in our memory, with approximately 20,000 individuals either deceased or unaccounted for. 
In Japan, earthquakes are the predominant form of natural disaster, having culminated in a substantial loss of life over time.
Such catastrophic events have continuously underscored the urgency of improving our understanding and predictive capabilities of earthquakes to better safeguard human societies. 
Given the unyielding and unpredictable nature of earthquakes, the need for precise prediction of seismic intensity distributions with precision is paramount. 
Such predictions can significantly aid in risk assessment, strengthening building structures, and planning strategic evacuations, thereby minimizing the devastating impacts of future seismic events.

\begin{figure}[t]
    \centering
    \includegraphics[width = \linewidth]{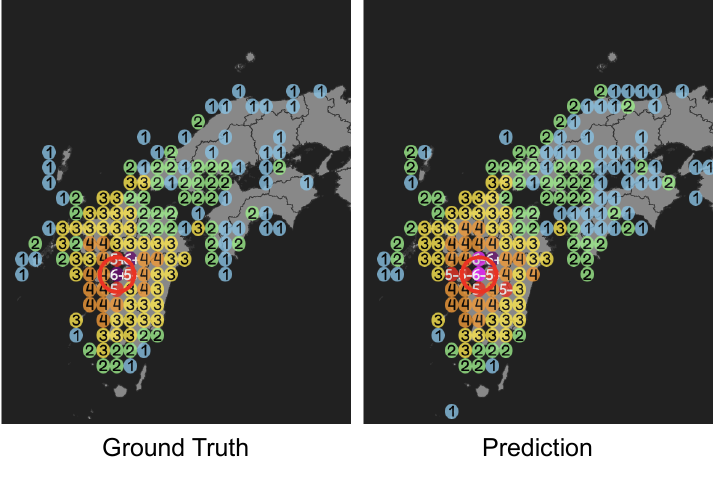}
    \caption{Comparison of seismic intensity distributions for a JMA-magnitude-6.4 earthquake that occurred at approximately 00:03 am JST, April 15, 2016, showing both the ground truth and our prediction. The hypocenter of the event is located at latitude $32^\circ42.0'$N, longitude $130^\circ°46.6'$E, and a depth $7$ km. Regions near the epicenter are cropped for better visualization.}
    \label{fig:earthquake_prediction_example}
\end{figure}

Conventionally, Ground Motion Prediction Equations~(GMPEs) have been the cornerstone of seismic intensity prediction methods~\cite{douglas2008survey, morikawa2013new}. 
GMPEs were developed primarily based on just two key predictive variables: magnitude and distance. 
Despite significant advancements in computer capabilities, enabling facile simulations of three-dimensional wave propagation, the simplistic GMPEs continue to be employed for seismic intensity predictions. 
Their enduring popularity can be attributed to their robustness, high reliability, ease of implementation, and low computational cost~\cite{DOUGLAS2016203}.

However, their effectiveness is limited by reliance on detailed geological data and an inherent limitation in expressive power.
GMPEs, based largely on empirical rules, incorporate complex geological variables such as the velocity amplification factor. 
As a result, GMPEs often struggle to accurately capture seismic intensity distributions influenced by unique local underground structures, such as abnormal seismic intensity distributions.
An abnormal seismic intensity distribution is a phenomenon in which an area farther from the hypocenter shakes more than a nearby area due to the influence of the subsurface structures.
GMPEs lack expressiveness because of their simple regression equations, making it difficult to reproduce the patterns of abnormal seismic intensity distributions.

In this study, we propose a linear regression model for predicting seismic intensity distributions.
The proposed model directly predicts seismic intensity distribution using the location, depth, and magnitude of the epicenter as inputs.
In contrast to GMPEs, The proposed model does not require geographic information or complex equation assumptions.
It only learns the pattern of seismic intensity distribution from past earthquake data and does not make any assumptions or set parameters for soil conditions, plate positions, or equations.

As shown in \cref{fig:earthquake_prediction_example}, the proposed model accurately predicts seismic intensity distribution, with an improved correlation coefficient of +0.06 compared to the conventional GMPEs.
Additionally, the proposed model demonstrated greater expressiveness than GMPEs, accurately predicting even abnormal seismic intensity distributions.
The proposed model does not require geographic information and complex equation assumptions, thereby successfully addressing the limitations of GMPEs.

This study serves as an extension of our previous research~\cite{ours}. 
In our prior work, we established both classification and regression models and further developed a hybrid model that intricately combines these approaches. 
Our previous model achieved a correlation coefficient of 0.78.
Our new contributions are threefold: the extension and publication of the dataset, enhancement of the hybrid model, and implementation and comparative experiments against conventional GMPEs.

In this research, we have expanded the dataset to include seismic intensity data from earthquakes in the vicinity of Japan spanning the years from 1997 to 2020. 
This dataset now includes a comprehensive collection of 1,857 instances of seismic intensity data for earthquakes with magnitudes of 5.0 or higher, sourced from the Japan Meteorological Agency. 
To enhance accessibility and promote collaborative research, we have published this dataset, ensuring open access for academic and research purposes.

Subsequently, we retrained our classification and regression models using this expanded dataset to enhance the models' learning from a broader and more diverse range of seismic scenarios. 
For the classification model, our retraining process involved refining the model to more effectively distinguish between different levels of seismic intensities. 
In contrast, the regression model was designed to more precisely predict the exact values of seismic intensities. 
The hybrid model, a key component of our study, was redesigned to simplify the equation.
% In our previous work, we combined the outputs of the classification and regression models using the following formula: the intensity $ I =I_{\text{Regression}} - \alpha \times (1 - I_{\text{Classification}}) $. 
% In this study, we adopted a simpler and more interpretable approach: the intensity $ I $ is defined as $ I_{\text{Regression}} $ where $ I_{\text{Classification}} > 0 $; otherwise, it is set to $ 0 $.
This modification aims to streamline the model's implementation and enhance interpretability, facilitating a clearer understanding of the respective contributions of the classification and regression models to the final prediction.

Moreover, we have implemented one of the conventional GMPEs and compared it to the proposed model.
A comprehensive comparative evaluation of the proposed model against conventional GMPEs revealed that the proposed model outperforms these standard approaches in both qualitative and quantitative aspects. 

The primary contributions of this paper are as follows:
\begin{itemize}
    \item We propose linear regression models, which are not limited by geological information constraints, such as underground structures and soil conditions, enabling accurate prediction of seismic intensity distributions.
    \item We demonstrate the model's effectiveness in predicting not only general seismic patterns but also the more challenging abnormal seismic intensity distributions.
    \item We conduct a comprehensive comparative analysis with GMPEs to show the superior performance of our approach.

\end{itemize}
The source code and datasets for the proposed model are available at \href{https://github.com/hogehoge}{https://github.com/KoyuMizutani/Earthquake}.

\section{Related Works}
Earthquake prediction research predominantly bifurcates into two distinct directions~\cite{Reddy2023}. 
The first direction aims to prognosticate the precise location and magnitude of imminent earthquakes. 
Specifically, using historical seismic data and geological surveys, researchers aim to forecast the location, magnitude, and potential impact of forthcoming earthquakes.
%Historically, humans have been intrigued by possible predictors, such as the unusual behaviors exhibited by animals before seismic events~\cite{Alarifi2012}. 
Numerous potential earthquake precursor phenomena have been documented, including variations in electric and magnetic fields, anomalous gas emissions, groundwater level fluctuations, temperature shifts, surface deformations, and seismic activity~\cite{Cicerone2009, Abri2022}.
In modern research, these phenomena, combined with historical seismic data, are exploited using a gamut of artificial intelligence methodologies ranging from rule-based systems to contemporary machine learning and deep learning frameworks~\cite{banna2020, Abebe2023, Hoque2018, Hoque2020, Asim2017, Asim2017-2, Muhammad2023, Salam2021, Cekim2021, Rasel2019, Mignan2020}.

The second direction is centered around the instantaneous detection of preliminary tremors preceding a significant seismic event, subsequently estimating both the scale of the impending earthquake and the potential aftershock magnitude. 
Such systems, known as Earthquake Early Warning~(EEW) systems, serve to swiftly detect seismic activities, project tremor intensities, and provide timely alerts. 
An EEW system generally encapsulates four critical phases: detection of a seismic event and its localization, magnitude estimation, prediction of seismic intensity distributions, and deciding whether to activate an alarm or not~\cite{Cremen2020}.
In the realm of EEW, machine learning techniques play an instrumental role in forecasting the magnitude, location, and scale of seismic events~\cite{Apriani2021, Bilal2022}.

Ground Motion Prediction Equations (GMPEs) have traditionally been fundamental tools for predicting seismic intensity~\cite{douglas2008survey, morikawa2013new}. 
Originally developed focusing mainly on two predictors—magnitude and distance—GMPEs remain in widespread use even though advancements in computational capabilities have facilitated more complex simulations, such as three-dimensional wave propagation. 
The persistent application of GMPEs owes to their proven robustness, reliability, ease of use, and minimal computational demands~\cite{DOUGLAS2016203}. 
Predominantly empirical, GMPEs incorporate critical geological factors such as the velocity amplification factor.

However, despite their widespread use, GMPEs have several inherent limitations. 
First, their lack of adaptability is a significant concern. GMPEs are traditionally formulated based on predefined assumptions derived from historical earthquake data. 
These assumptions, though robust, might not always apply universally to each seismic event. This characteristic renders GMPEs somewhat rigid, limiting their effectiveness in novel or unusual earthquake scenarios. 
Second, there is an over-reliance on geological information in these conventional methods. 
They depend on detailed geological data, which can be a drawback in areas where such information is lacking or outdated, leading to potentially inaccurate earthquake impact predictions. 
Third, the ability of GMPEs to capture complex patterns, which are frequently observed in earthquakes and their effects, is another area where they fall short. 
Such complexities in seismic activity often extend beyond the linear predictions provided by these models, underscoring the need for more sophisticated approaches in earthquake intensity prediction.
Recent advancements have seen a foray into machine learning to bolster the capabilities of GMPEs~\cite{derras2012adapting, Hybrid}. 
While these integrated approaches are promising, they typically combine machine learning techniques with conventional GMPEs, inheriting their shortcomings.

We aim to predict seismic intensity distribution in a data-driven manner, without making complex equation assumptions. 
The proposed model is independent of geographic information and provides more accurate seismic intensity distribution predictions than conventional GMPEs.

\section{Preliminary}
Based on \cite{recipe}, we implemented a conventional method of GMPEs as the baseline for comparison. The peak ground velocity on the engineering bedrock~($PGV_b$) is expressed using moment magnitude~($M_{\mathrm{w}}$), hypocenter depth~($D$), and fault shortest distance~($X$) as follows\cite{si1999}.
\begin{align}
  \begin{split}
    \log P G V_b = &0.58 M_{\mathrm{w}}+0.0038 D-1.29 \\
    &-\log \left(X+0.0028 \cdot 10^{0.50 M_{\mathrm{w}}}\right)-0.002 X .
  \end{split}
\end{align}
In this study, the value of $X$ was approximated as the distance between the hypocenter and the target point. The distance was calculated using the haversine formula, a method for calculating the great-circle distance between two points on a sphere.

Moment magnitude~($M_{\mathrm{w}}$) and JMA magnitude~($M_{\text{JMA}}$) are related via seismic moment~($M_0$) as follows\cite{Kanamori1977, TAKEMURA1990}.
\begin{align}
    M_{\mathrm{w}} &= (\log M_0 - 9.1)/1.5 , \\
    \log M_0 &= 1.17 \cdot M_{\text{JMA}} + 10.72 , \\
    \therefore M_{\mathrm{w}} &= 0.78 \cdot M_{\text{JMA}} + 1.08 .
\end{align}
The maximum velocity amplification factor~($amp$) is calculated using the average S-wave velocity at 30m below the surface~($AVS30$)\cite{FUJIMOTO2006relationship}. The $AVS30$ values were downloaded from Japan Seismic Hazard Information Station\footnote{\url{https://www.j-shis.bosai.go.jp/}}\cite{wakamatsu20131, wakamatsu20132, matsuoka_wakamatsu_2008, FUJIMOTO2006relationship}.
The minimum non-zero $AVS30$ was obtained so that $amp$ would be the maximum in each cell on the map.
\begin{align}
    \log (a m p)=2.367-0.852 \cdot \log A V S 30 \pm 0.166 .
\end{align}
The peak ground velocity on the engineering bedrock~($PGV_b$) is multiplied by the amplification factor~($amp$) to obtain the peak ground velocity~($PGV$) on the ground surface.
\begin{align}
    P G V=a m p \cdot P G V_{\mathrm{b}}.
\end{align}
The instrumental seismic intensity~($I$) is calculated using $PGV$ as follows\cite{Fujimoto2005}.
\begin{align}
  I = 
  \begin{cases}
    2.165 + 2.262 \cdot \log (P G V), & \quad (I < 4) \\
    \vspace{-3mm} \\
    \begin{aligned}
      2.002 + 2.603 \cdot \log (P G V) \\
      - 0.213 \cdot \{\log (P G V)\}^2.
    \end{aligned} & \quad (4 \leq I) 
  \end{cases}
\end{align}

\section{Methodology}
This study aims to enhance the prediction accuracy of seismic intensity distributions using linear regression models. 
We have developed three types of models: classification, regression, and hybrid models.
These models represent advancements over our initial versions as described in our previous work~\cite{ours}.

The proposed model was specifically designed to prioritize simplicity and to predict accurately without the need for geographical data. 
This focus was a key aspect of our approach, addressing the shortcomings of GMPEs and maintaining computational cost-effectiveness. 
We developed and trained a linear regression model and a classification model to predict seismic intensity distribution.
Additionally, we adopted a hybrid model that combines the regression and classification models to harness the strengths of both approaches, providing more accurate predictions of seismic intensity distributions.

\subsection{Dataset}
The dataset comprises seismic intensity data from earthquakes that occurred in Japan between 1997 and 2020. The dataset contains 1,857 instances of seismic intensity data for earthquakes with a magnitude of 5.0 or greater, obtained from the Japan Meteorological Agency~(JMA)\footnote{\url{https://www.data.jma.go.jp/eqev/data/bulletin/shindo_e.html}}. 
The dataset includes the latitude, longitude, and depth of the hypocenter, the JMA magnitude ($M_{\text{JMA}}$), and the recorded instrumental seismic intensity at various observation stations. 
We divided the dataset into 1,455 training, 227 validation, and 175 test samples.

\begin{figure}[t]
    \centering
    \includegraphics[width = 0.9\linewidth]{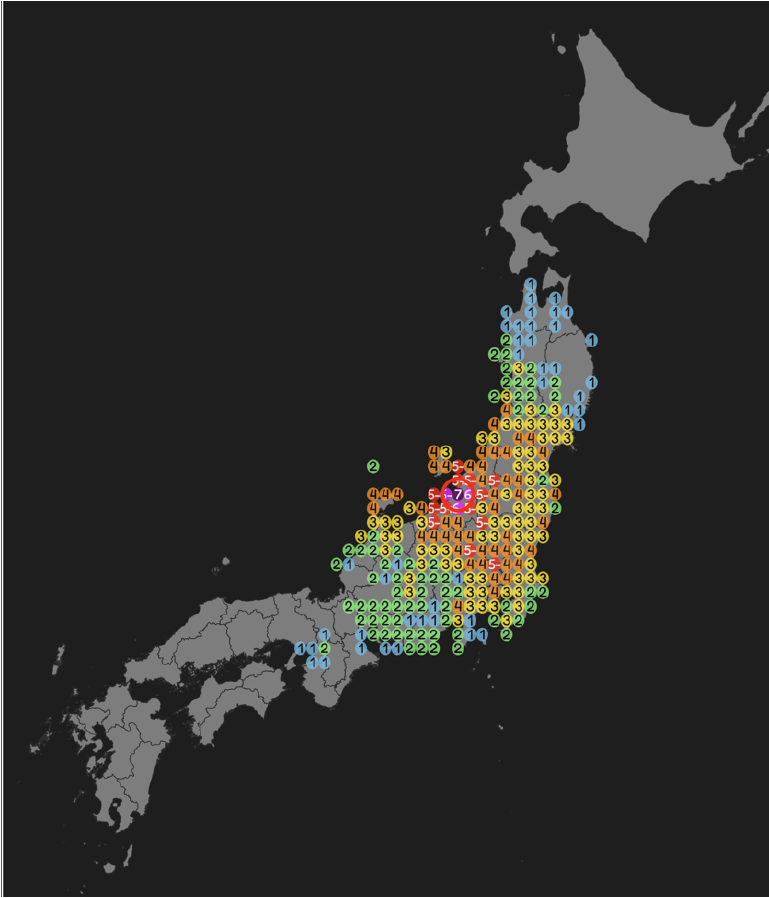}
    \caption{Seismic intensity distribution of the 2004 Chuetsu earthquake. Observed seismic intensities are represented in each grid cell. The epicenter is marked with a red circle.}
    \label{fig:earthquake_distribution_example}
\end{figure}

The seismic intensities at each observation point are compiled into a map format based on the coordinates of the observation stations.
The seismic intensity distribution data are transformed into a grid format, divided into $64\times64$ cells. 
This grid corresponds to a section of the Mercator map, specifically the rectangular region spanning from $30^\circ$N to $46^\circ$N, and $128^\circ$E to $146^\circ$E which is divided into $64\times64$ cells.
The seismic intensity recorded at each observation station is assigned to the corresponding cell, based on the station's coordinates.
In cells with multiple observation points, the highest recorded seismic intensity among them is used. For cells without an observation station, the seismic intensity is designated as 0.

The seismic intensity values in the dataset are provided in terms of instrumental seismic intensity~(continuous values).
However, when drawn on a map, the seismic intensity is represented in 10 degrees based on the JMA Seismic Intensity Scale\footnote{\url{https://www.jma.go.jp/jma/en/Activities/inttable.html}}.
\cref{fig:earthquake_distribution_example} shows an example of the seismic intensity distribution for a specific earthquake processed as described above.

\subsection{Inputs}
The proposed model uses the earthquake's hypocenter latitude, longitude, depth, and the JMA magnitude as input features.
To enable the model to process two-dimensional geographical data, these inputs are formatted into a $64 \times 64$ matrix. 
This representation is ultimately flattened before being fed into the network.
All cells in the matrix are initialized to zero, after which
the hypocenter's depth and magnitude values are assigned to their respective cells. 

\begin{figure}[t]%[htbp]
\centering
\includegraphics[width = 0.6\linewidth]{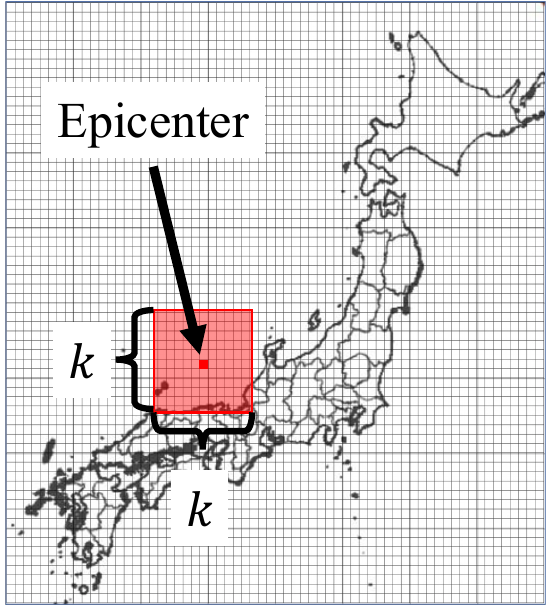}
\caption{Illustration of the input format. The depth and magnitude values are assigned across $k \times k$ cells, centering around the hypocenter's cell. All non-assigned cells retain 0.}
\label{fig:input_format}
% \vspace{-0.19cm}
\end{figure}

Upon entering the network, the $64 \times 64$ matrix is flattened.
This transformation has a risk of obscuring the inherent geographic relationships between cells. 
To maintain this spatial information and enhance the model's accuracy, depth and magnitude values are propagated over a $k\times k$ cellular area centered around the epicenter. 
Each cell within this $k\times k$ area is assigned the same depth and magnitude values. 
The optimal size of $k$ is empirically derived to maximize performance indicators. 
\cref{fig:input_format} represents the input format articulated above. 

Inputs are structured as a three-dimensional array, with depth and magnitude metrics embedded along the z-axis within the specified $k \times k$ cellular space. 
The magnitude values undergo an exponential or power transformation to ensure the model's interpretability. 

\subsection{Classification and Regression Model}
\begin{figure*}[t]
    \centering
    \includegraphics[width = \linewidth]{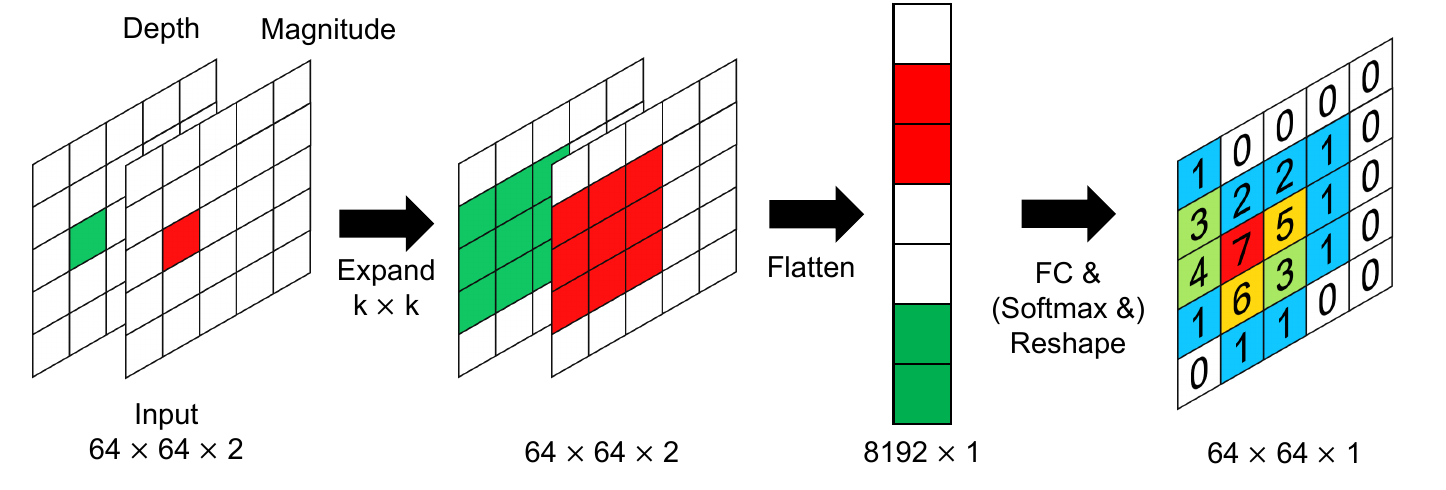}
    \caption{Overview of the architecture used for both the classification and regression models. $64\times 64$ cells are flattened and input as a vector to the network. For the classification model, the network outputs the probabilities of seismic intensity classes at each cell. The regression model outputs continuous instrumental seismic intensity values.}
    \label{fig:cls_reg_models}
    % \vspace{-0.19cm}
\end{figure*}

Both the classification and regression models in this study consist of a single fully connected layer.
\cref{fig:cls_reg_models} illustrates the architecture shared by both models.
The data from the $64\times 64$ cells, which include the depth and the magnitude of an earthquake, are flattened for input into the models.

In the classification model, seismic intensity data are converted to the JMA Seismic Intensity Scale format for processing.
In this model, a softmax function follows the fully connected layer, generating probabilities for various seismic intensity classes corresponding to each cell.
These probabilities are then reshaped to match the original geographic distribution.

The key difference between the classification and regression models lies in the function utilized at the output layer. 
The classification model employs a softmax function to yield probabilities for each seismic intensity class, whereas the regression model outputs continuous values representing instrumental seismic intensity directly.
Hence, the regression model behaves as a simple linear regressor.

Both models are designed to capture geographic relationships among cells by making simultaneous predictions for each cell. 
This approach enables the models to effectively consider spatial dependencies within the seismic intensity distribution.

In training, we set a batch size to 128, epochs, and a learning rate to 0.1. 
We employed the Adam optimizer and used cross-entropy loss for the classification model and Mean Squared Error (MSE) for the regression model. 
The values of $k$ were set at 17 for classification and 5 for regression.
The optimal $k$ was empirically determined to maximize the correlation coefficient between the ground truth and the predicted outcomes.
Weights were adjusted appropriately to accommodate imbalanced data.

\subsection{Hybrid Model}
The hybrid model is designed to combine the strengths of the regression and classification models. 
The regression model, which provides a continuous output, is leveraged for its proficiency in making precise estimates close to the earthquake epicenter. 
Conversely, the classification model, which outputs discrete class probabilities, is used for its ability to more precisely predict whether seismic intensity is greater than zero. 
Integrating these two models in the hybrid framework aims to enhance the accuracy and reliability of predictions over a wider spectrum of seismic intensities and geographical zones.
The output of the hybrid model is computed as follows:
\[
I = 
\begin{cases} 
I_{\text{Regression}} & \text{where } I_{\text{Classification}} > 0, \\
0 & \text{otherwise.}
\end{cases}
\]
%The motivation for this design is to leverage the regression model's ability to make accurate predictions close to the earthquake's epicenter, while also utilizing the classification model's effectiveness in determining whether the seismic intensity is greater than zero.

\section{Results}

\subsection{Evaluation Metrics}
We use three metrics for the quantitative evaluation of the proposed models: the correlation coefficient~($r$), F1 score, and Matthews Correlation Coefficient (MCC).
The predicted seismic intensities are first rounded to the nearest instrumental class value on the JMA Seismic Intensity Scale before calculating these metrics. 
The choice of these metrics aims to provide a balanced assessment of the proposed models' performance in predicting seismic intensities.

\begin{itemize}
    \item \textbf{Correlation Coefficient~($r$):} Correlation coefficient measures the linear relationship between the rounded predicted and actual seismic intensities. It helps to understand how well the model captures the underlying trend in seismic intensities across different classes.
    
    \item \textbf{F1 Score:} F1 score is calculated as a binary classification metric, determining whether the predicted intensity is zero or greater than zero. It provides a balanced measure of a model's performance when classes are imbalanced, which is often the case in seismic intensity predictions where higher intensities are rare.
    
    \item \textbf{Matthews Correlation Coefficient (MCC):} MCC is calculated considering the rounded predictions as a multi-class classification problem. This metric offers a comprehensive measure of classification quality, taking into account true and false positives and negatives for each class on the JMA scale. The MCC generates high scores only if the predictions are good for all classes~\cite{chicco2020advantages}.
\end{itemize}

\subsection{Qualitative Evaluation}
\begin{figure*}[t]
    \centering
    \includegraphics[width = \linewidth]{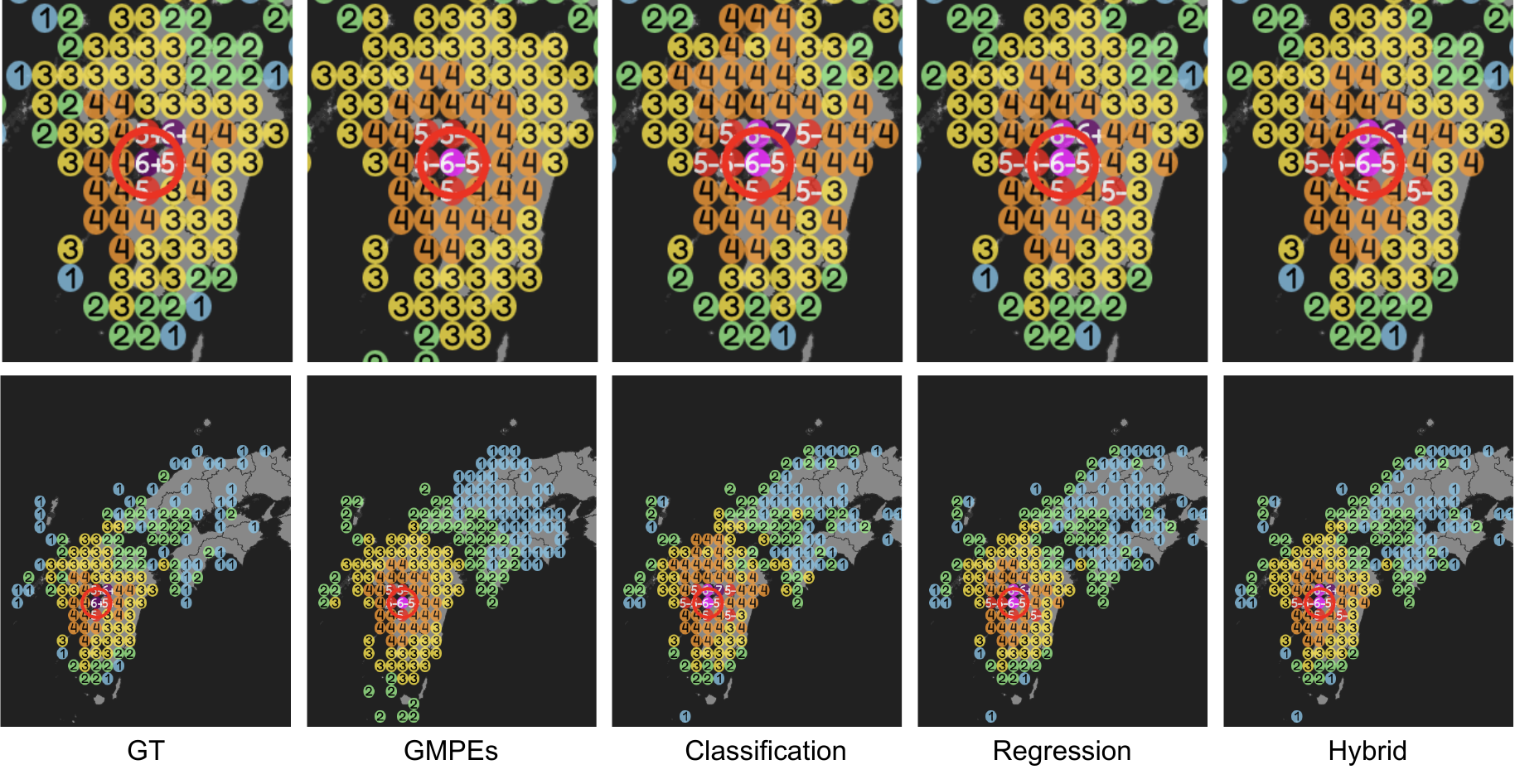}
    \caption{
    Example comparison of seismic intensity distributions for a JMA-magnitude-6.4 earthquake occurred at approximately 00:03 JST, April 15, 2016: ground truth, GMPEs, regression, classification, and hybrid models. The hypocenter of the event is located at latitude $32^\circ42.0'$N, longitude $130^\circ°46.6'$E, and depth $7$ km. The regions near the epicenter are cropped for better visualization.}
    \label{fig:qualitative_results_kumamoto}
\end{figure*}

\begin{figure*}[t]
    \centering
    \includegraphics[width = \linewidth]{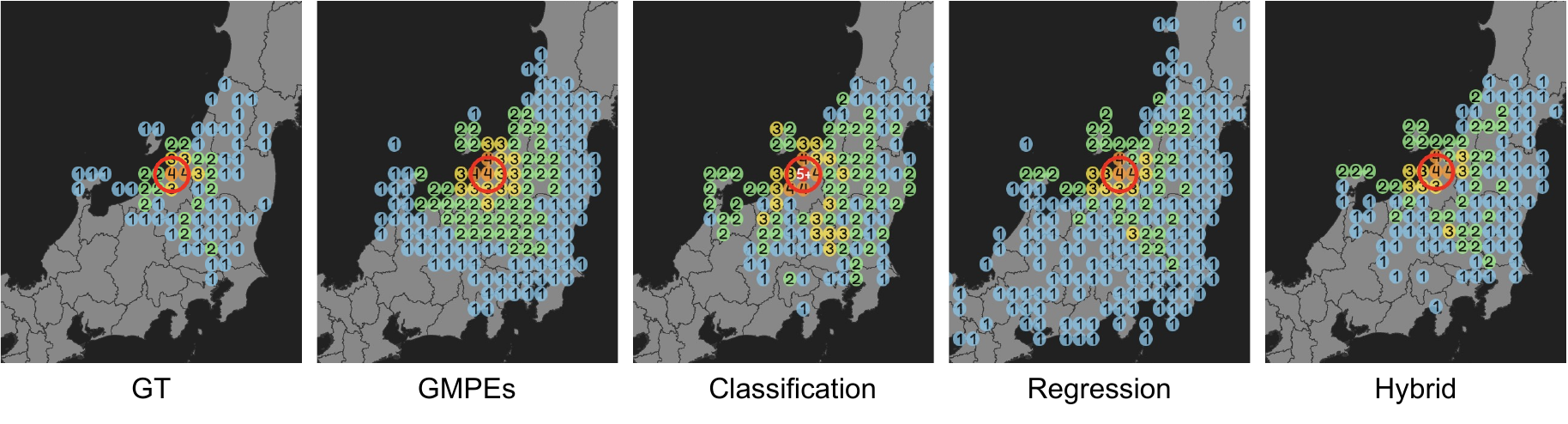}
    \caption{%Example comparison of seismic intensity distributions for the relatively weaker earthquake~(JMA Magnitude 5.0). The epicenter is located in Niigata prefecture.
    Example comparison of seismic intensity distributions for the relatively weaker earthquake~(JMA Magnitude 5.0) which occurred at approximately 21:44 JST, October 23, 2004. %The epicenter is located in Niigata prefecture.
    The hypocenter is located at latitude $37^\circ16.3'$N, longitude $138^\circ°56.5'$E, and depth $15$~km.}
    \label{fig:qualitative_results_nagano}
\end{figure*}

\cref{fig:qualitative_results_kumamoto,fig:qualitative_results_nagano} present comparative evaluations of seismic intensity distributions for two distinct earthquakes: 
%the strong 2016 Kumamoto Earthquake and another event of relatively lower magnitude. 
a strong earthquake in the 2016 Kumamoto earthquake~(Kumamoto earthquake sequence) and another event of relatively lower magnitude. 
The ground truth for each provides the observed seismic intensities and the GMPEs, regression, classification, and hybrid models each offer their predictions.

Upon visually inspecting the Kumamoto earthquake sequence as shown in ~\cref{fig:qualitative_results_kumamoto}, it is evident that the GMPEs produce a uniform prediction, failing to accurately represent the complex intensity distribution of the actual event.
The regression model excels in predicting the exact intensity values, notably in areas closer to the epicenter. 
It excels at capturing the actual values but falters in accurately predicting the range of the intensity distribution. 
Conversely, the classification model, while tending to overestimate the intensities, excels at predicting the range of the seismic intensity distribution, making it a valuable complement to the regression model. 
The hybrid model, effectively leveraging the strengths of both the regression and classification models, offers a seismic intensity distribution prediction that is the most closely aligned with the ground truth.

Similar patterns are observed for the weaker earthquake as shown in~\cref{fig:qualitative_results_nagano}. 
The GMPEs' prediction is consistently simplistic, lacking the nuanced variations that are evident in the actual seismic intensities. 
The regression model accurately predicts the values but lacks precision in capturing the range, while the classification model does the opposite, capturing the range but overestimating the values. 
In conclusion, the hybrid model emerges as the most reliable option, offering well-rounded and accurate predictions, thus confirming its applicability and effectiveness across earthquakes of varying magnitudes and conditions.

\subsection{Quantitative Evaluation}

We evaluated the proposed model using three metrics: correlation coefficient, F1 score, and MCC. 
The results are summarized in Table~\ref{tab:quantitative_metrics}.

\begin{table}[t]
    \centering  
    \normalsize
    \caption{Quantitative metrics for each model.}
    \begin{tabular}{lcccc} \toprule
        Model & $r$ & F1 Score & MCC \\
        \midrule
        GMPEs & 0.76  & 0.64  & 0.51 \\
        Previous Model & 0.78 & 0.61 & - \\
        Classification & 0.75  & 0.70  & 0.52 \\
        Regression & 0.77  & 0.54  & 0.48 \\
        Hybrid & \bf{0.82} & \bf{0.72}  & \bf{0.59} \\
        \bottomrule
    \end{tabular}
    \label{tab:quantitative_metrics}
\end{table}

The regression model exhibited strong performance in terms of the correlation coefficient, indicating its capability to capture the overall trend of seismic intensity. 
On the other hand, the classification model excelled in the F1 score, demonstrating its proficiency in identifying whether the seismic intensity is zero or greater than zero. 
Remarkably, the hybrid model achieved the highest scores across all metrics.
This suggests that the hybrid model is well-suited for capturing both the general trend and specific nuances in seismic intensity prediction.

\subsection{Distribution of Model Predictions}

\begin{figure*}[t]
    \centering
    \includegraphics[width = \linewidth]{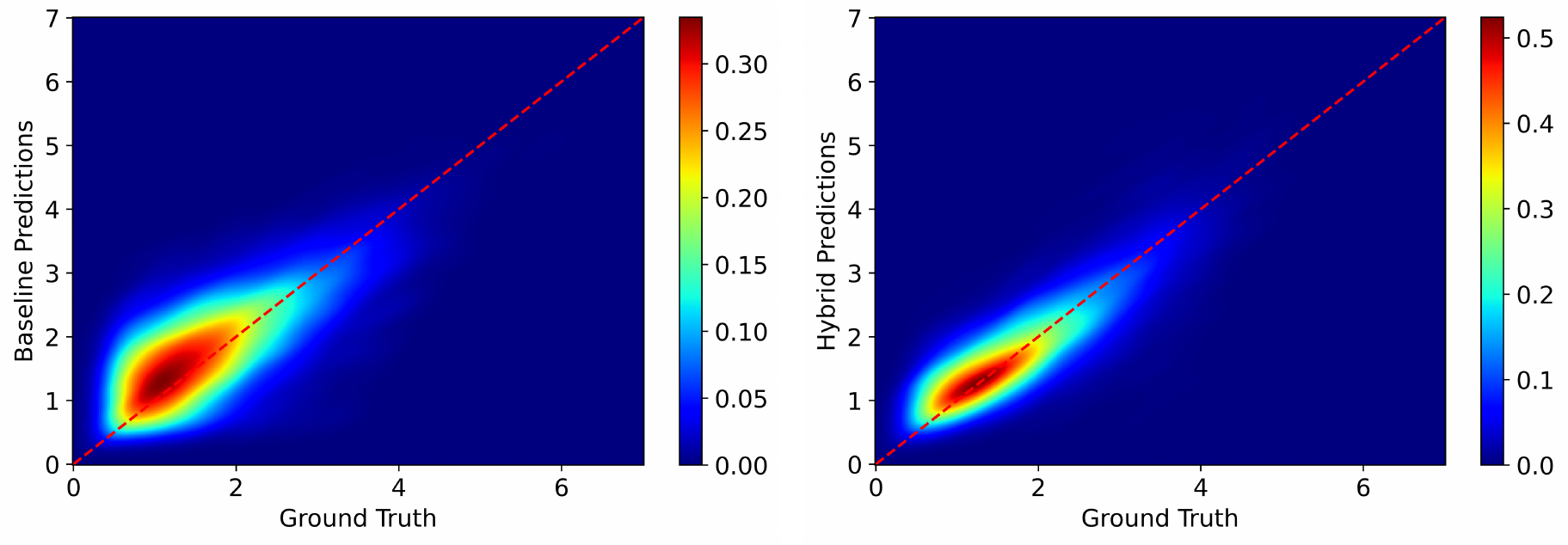}
    \caption{Kernel density estimation of the instrumental seismic intensity predictions compared to the ground truth. Left: GMPEs. Right: proposed model~(hybrid model).}
    \label{fig:heatmap}
\end{figure*}

\cref{fig:heatmap} visually illustrates the comparison of kernel density estimates for the predicted instrumental seismic intensities between the GMPEs and the hybrid models. 
The distribution is concentrated between seismic intensity 0.5 and 2, with the frequency of occurrence decreasing as the seismic intensity increases. Values between seismic intensity 0 and 0.5 do not appear because instrumental seismic intensities below 0.5 are rounded to 0.

\cref{fig:heatmap} shows that the proposed model's predictions are more concentrated around the ground truth, indicating sharper and more accurate predictions. In contrast, the GMPEs' predictions are dispersed, revealing their limited capability to match the actual seismic intensities closely.
This concentration in the prediction distribution of the proposed model near the ground truth signifies its superior performance, providing a clearer, more confident prediction of seismic intensities. 
% Such a stark contrast in distribution reaffirms the efficacy of the proposed model over the GMPEs.

\subsection{Predicting Abnormal Seismic Intensity Distributions}
The ability of the proposed model to forecast abnormal seismic intensity distributions is noteworthy. 
Deep earthquakes sometimes manifest an abnormal seismic intensity distribution, where the fore-arc motions are more potent than those in the back-arc~\cite{iwakiri2011improvement}. 
This inversion of expected seismic intensity, where areas farther from the hypocenter experience more substantial tremors, has been challenging for conventional prediction methods such as GMPEs. 
This is because such methods demand a nuanced comprehension of local subsurface structures, a feat not always achievable. 

\begin{figure*}[t]
    \centering
    \includegraphics[width = \linewidth]{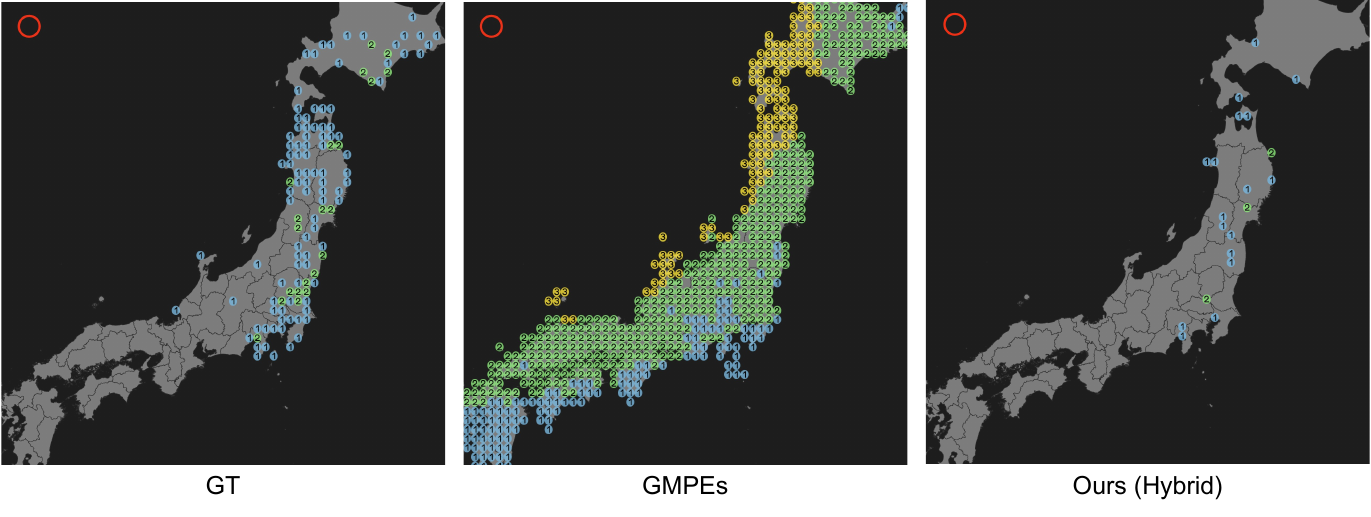}
    \caption{%Comparison of observed and predicted seismic intensity distributions for an earthquake with an abnormal distribution~(Magnitude 7.1). The hypocenter is situated near Vladivostok, and the seismic intensity spreads towards the east of the Tohoku region. The proposed model accurately captures this abnormal distribution pattern.
    Comparison of observed and predicted seismic intensity distributions for an earthquake with an abnormal distribution~(JMA Magnitude 7.1) which occurred at approximately 22:10 JST, April 8, 1999. The hypocenter is located at latitude $43^\circ33.0'$N, longitude $130^\circ°59.4'$E, and depth $633$~km. The epicenter is located near Vladivostok, and the seismic intensity spreads towards the east of the Tohoku region. The proposed model accurately captures this abnormal distribution pattern.}
    \label{fig:abnormal}
\end{figure*}

However, the proposed model can predict these abnormal distributions. 
For example, \cref{fig:abnormal} elucidates the observed and predicted outcomes of an abnormal seismic intensity distribution. 
In this example, with the hypocenter located in the vicinity of Vladivostok, the tremors are distributed toward the east of the Tohoku region. 
The proposed model successfully anticipates this distribution.
This successful prediction indicates that the proposed model possesses greater versatility and expressiveness of the proposed model compared to GMPEs. 
Moreover, the network appears to incorporate the spatial relationships of tectonic plates and other subterranean structures during its learning process from historical earthquake data.

\section{Discussions}
\subsection{Limitations}
The proposed model, rooted in a data-driven approach, possesses several inherent limitations. 
First, its accuracy is intrinsically tied to the volume of a training dataset. 
A paucity of adequate data, especially from underrepresented regions or infrequent seismic occurrences, might adversely affect its predictive prowess. 
Second, similar to many end-to-end networks, the proposed model faces interpretability challenges. 
The model's black-box nature makes it difficult to discern the reasoning behind specific predictions, which could be problematic where clarity is essential for certain applications or in building user trust. 
Despite these promising results in predicting seismic intensity distributions, one should exercise caution and account for these constraints when considering its application in real-world settings or diverse geographical regions.

The current architecture of the proposed model demonstrates its potential for predicting seismic intensity distributions using neural networks. 
Increasing the number of layers and building neural networks could help in capturing more complex patterns present in the seismic data, which might lead to improved prediction accuracy.
Nevertheless, a deeper network architecture is not always essential, as the proposed model outperforms GMPEs even with just a single fully connected layer.
Deeper network architectures can potentially have an overfitting problem, as well.

\subsection{Future Work}
The analysis of the predictive outcomes from the proposed model can also be utilized to identify regions with a higher susceptibility to earthquakes. 
Such insights can provide valuable information for urban planners and policymakers to prioritize safe infrastructure development in areas identified as high risk.

Real-time seismic intensity predictions, informed by timely and accurate data, could enhance early warning systems and facilitate prompt response measures.
Furthermore, integrating the proposed model with existing earthquake prediction or warning systems could enhance seismic predictions' overall accuracy and reliability. 
Collaborative efforts between different prediction systems and databases may lead to a more robust earthquake preparedness framework.

A noteworthy finding from our experiments is the model's potential ability to discern the characteristics of underground structures. 
Based on this observation, we hypothesize that seismic waves could be conceptualized as analogous to rays of light. 
By adopting this perspective, we believe using models similar to NeRF~\cite{nerf} for estimating densities at various coordinates beneath the Earth's surface is feasible. 
This approach would harness NeRF's capability to infer 3D structures, potentially enabling a more detailed and accurate reconstruction of subsurface formations based solely on seismic data.

\section{Conclusions}
In this study, we proposed linear regression models to predict seismic intensity distributions.
The proposed model predicts the seismic intensity for each location in the vicinity of Japan based on the earthquake's hypocenter latitude, longitude, depth, and magnitude.
In contrast to GMPEs, which are most commonly used to predict seismic intensity distribution, the proposed model does not require complex equation assumptions or geographic information and is trained in a data-driven manner.
Classification, regression, and a hybrid model of the two were created and trained with data from 1,857 earthquakes of magnitude 5 or greater that occurred near Japan between 1997 and 2020.
The proposed model outperformed conventional GMPEs in accuracy on three measures: correlation coefficient, F1 score, and MCC.
In addition, the proposed model provided accurate predictions for abnormal seismic intensity distributions, which are difficult to predict with conventional GMPEs.
This advancement represents a significant step forward in earthquake prediction, offering a more reliable tool for preparing and mitigating the impacts of these natural disasters.

%However, our approach has some limitations. 
%The reliance on data-driven methods inherently ties the accuracy of our model to the quantity and quality of a training dataset. 
%In addition, similar to many other end-to-end networks, the model's output is difficult to interpret.

%For future work, enhancing the model's complexity by integrating more layers or adopting advanced architectures holds the potential for even better predictive accuracy. 
%Moreover, expanding our approach to perform geographical risk analysis can offer invaluable insights into potential danger zones. 
%Pursuits into real-time prediction and integration of our model with existing systems can further extend its utility and societal impact.

\bibliographystyle{IEEEtran}
\bibliography{bib/main}

% \newpage

% \section{Biography Section}
% If you have an EPS/PDF photo (graphicx package needed), extra braces are
%  needed around the contents of the optional argument to biography to prevent
%  the LaTeX parser from getting confused when it sees the complicated
%  $\backslash${\tt{includegraphics}} command within an optional argument. (You can create
%  your own custom macro containing the $\backslash${\tt{includegraphics}} command to make things
%  simpler here.)
 
% \vspace{11pt}

% \bf{If you include a photo:}\vspace{-33pt}
% \begin{IEEEbiography}[{\includegraphics[width=1in,height=1.25in,clip,keepaspectratio]{fig/fig1}}]{Michael Shell}
% Use $\backslash${\tt{begin\{IEEEbiography\}}} and then for the 1st argument use $\backslash${\tt{includegraphics}} to declare and link the author photo.
% Use the author name as the 3rd argument followed by the biography text.
% \end{IEEEbiography}

\begin{IEEEbiography}[{\includegraphics[width=1in,height=1.25in,clip,keepaspectratio]{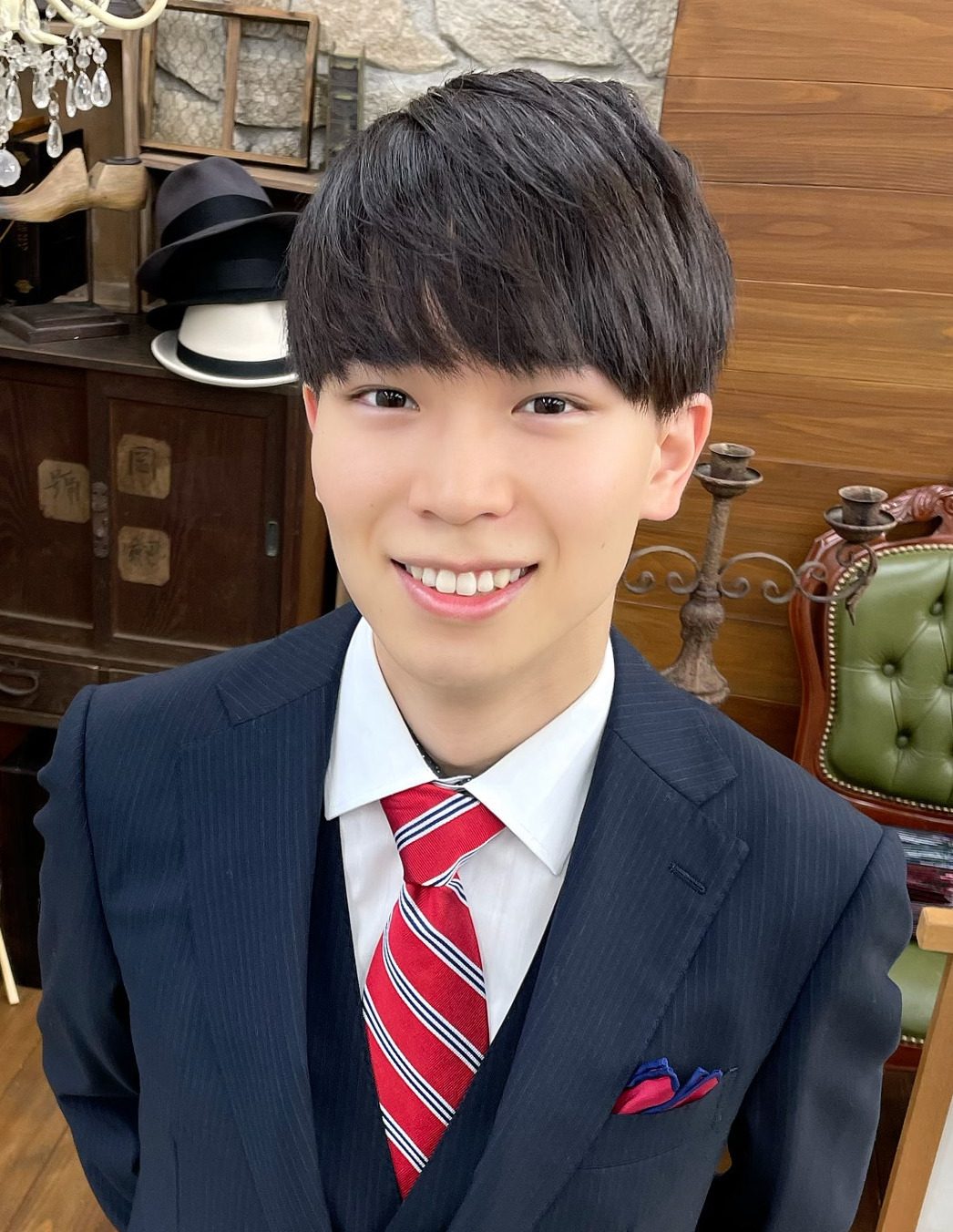}}]{Koyu Mizutani}
received his B.Eng. degree from The University of Tokyo.
He is currently a Master's student at the Department of Information and Communication Engineering, Graduate School of Information Science Technology, The University of Tokyo. His research topics are hazard assessment with machine learning and presentation slide analysis. His current research interests include 3D point cloud data, large language models, multimodal data analysis, and so on.
\end{IEEEbiography}

\begin{IEEEbiography}[{\includegraphics[width=1in,height=1.25in,clip,keepaspectratio]{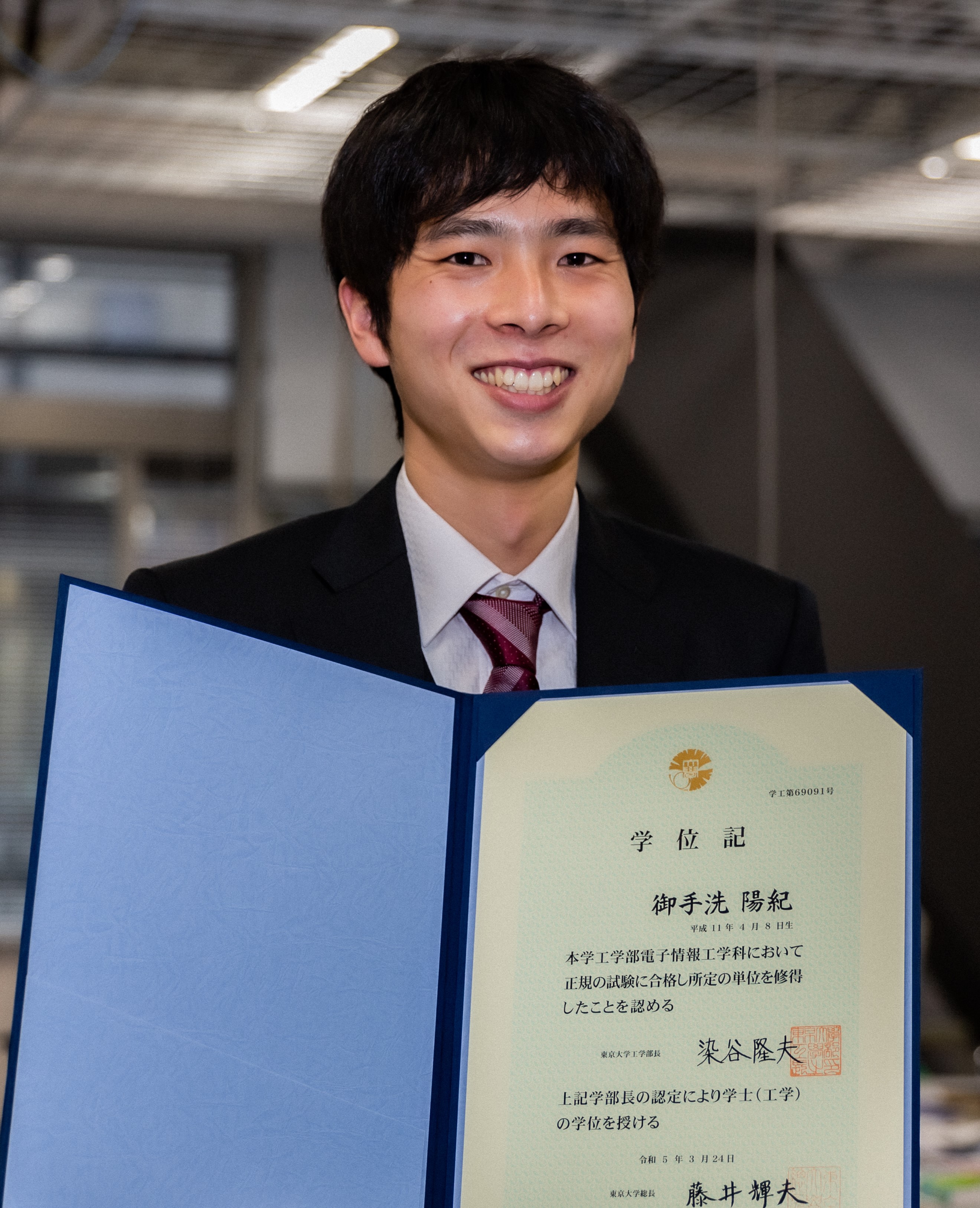}}]{Haruki Mitarai}
received his B.Eng. degree from The University of Tokyo.
He is currently a Master's student at the Department of Information and Communication Engineering, Graduate School of Information Science Technology, The University of Tokyo. 
His current research interests include quantum physics, relational databases, and the application of quantum computing to database systems.
\end{IEEEbiography}

\begin{IEEEbiography}[{\includegraphics[width=1in,height=1.25in,clip,keepaspectratio]{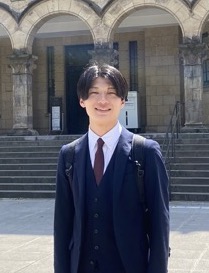}}]{Kakeru Miyazaki}
is currently a Master's student at the Graduate School of Interdisciplinary Information Studies, The University of Tokyo. His research topic is human-computer interaction.
\end{IEEEbiography}

\begin{IEEEbiography}[{\includegraphics[width=1in,height=1.25in,clip,keepaspectratio]{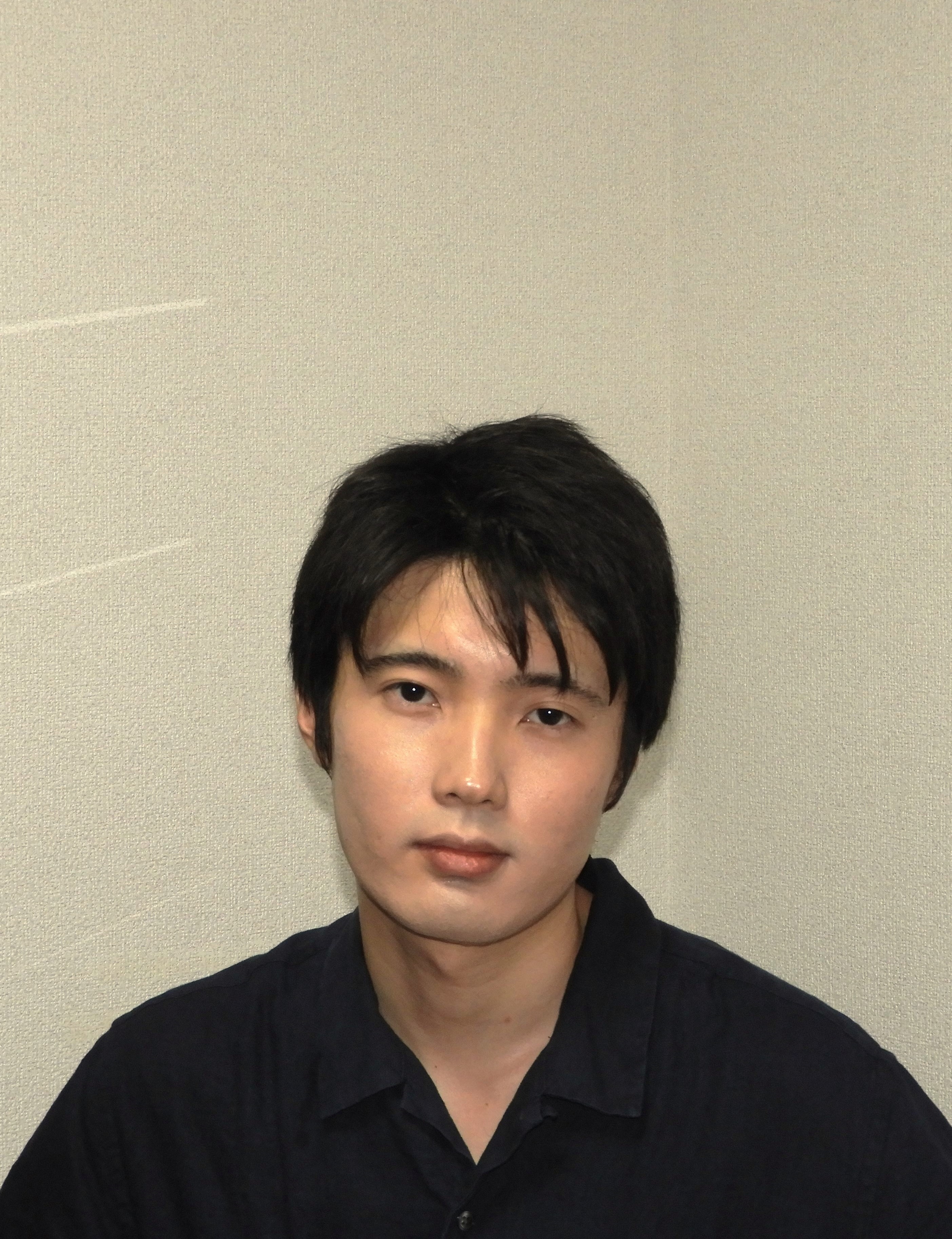}}]{Soichiro Kumano}
is currently a Ph.D. student at the Department of Information and Communication Engineering, Graduate School of Information Science Technology, The University of Tokyo. His research topics are pattern recognition, machine learning, and the theoretical understanding of deep learning.
\end{IEEEbiography}

\begin{IEEEbiography}[{\includegraphics[width=1in,height=1.25in,clip,keepaspectratio]{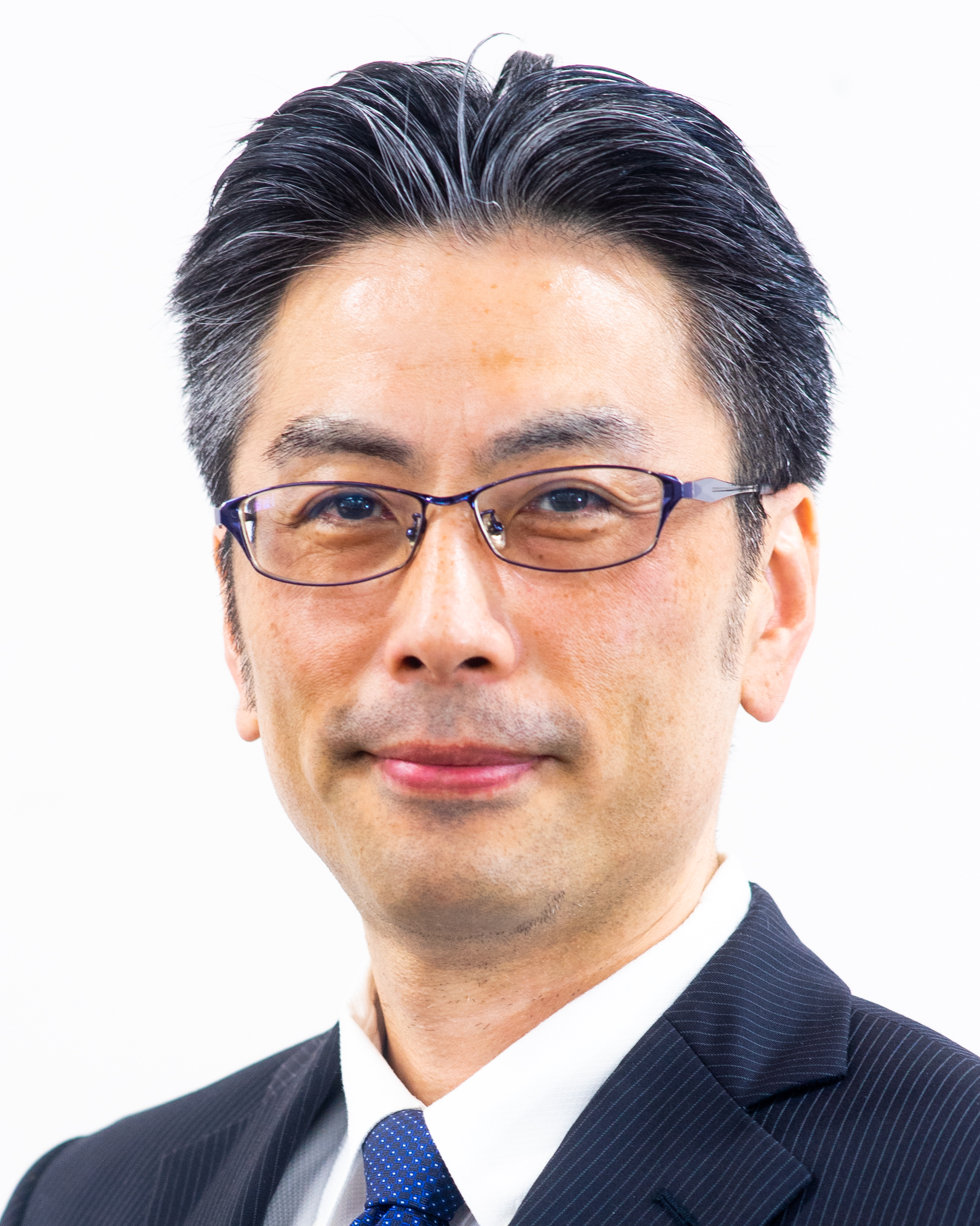}}]{Toshihiko Yamasaki}
received the Ph.D. degree from The University of Tokyo. He is currently a Professor at the Department 
of Information and Communication Engineering, Graduate School of InformationScience and Technology, The University of Tokyo. He was a JSPS Fellow for Research Abroad and a visiting scientist at Cornell University from Feb. 2011 to Feb.2013.His current research interests include attractiveness computing based on multimodal data analysis, pattern recognition, machine learning, and so on.
\end{IEEEbiography}

% \vspace{11pt}

% \bf{If you will not include a photo:}\vspace{-33pt}
% \begin{IEEEbiographynophoto}{John Doe}
% Use $\backslash${\tt{begin\{IEEEbiographynophoto\}}} and the author name as the argument followed by the biography text.
% \end{IEEEbiographynophoto}

\vfill

\end{document}